\pdfoutput=1

\documentclass[11pt]{article}

\usepackage[final]{acl}

\usepackage{times}
\usepackage{latexsym}

\usepackage[T1]{fontenc}

\usepackage[utf8]{inputenc}

\usepackage{microtype}

\usepackage{inconsolata}

\usepackage{graphicx}

\definecolor{OliveGreen}{rgb}{0,0.6,0}
\definecolor{CornellRed}{rgb}{0.7, 0.11, 0.11}

\usepackage{booktabs, multirow, makecell}
\usepackage{enumitem}
\usepackage{xcolor,colortbl}
\usepackage{amsmath}
\usepackage{amssymb}
\usepackage{pifont}
\title{DiVERT: \textbf{Di}stractor Generation with \textbf{V}ariational \textbf{E}rrors \\ \textbf{R}epresented as \textbf{T}ext for Math Multiple-choice Questions}

\author{
Nigel Fernandez\textnormal{\textsuperscript{1*}}, Alexander Scarlatos\textnormal{\textsuperscript{1*}}, Wanyong Feng\textnormal{\textsuperscript{1}} \\\bf Simon Woodhead\textnormal{\textsuperscript{2}}, Andrew Lan\textnormal{\textsuperscript{1}} \\
\textsuperscript{1}University of Massachusetts Amherst, \textsuperscript{2}Eedi \\
\texttt{\{nigel,ajscarlatos,wanyongfeng,andrewlan\}@cs.umass.edu}\\\texttt{simon.woodhead@eedi.co.uk}
}

\begin{document}
\maketitle

\def\thefootnote{*}\footnotetext{These authors contributed equally to this work.}\def\thefootnote{\arabic{footnote}}

\begin{abstract}
High-quality distractors are crucial to both the assessment and pedagogical value of multiple-choice questions (MCQs), where manually crafting ones that anticipate knowledge deficiencies or misconceptions among real students is difficult. Meanwhile, automated distractor generation, even with the help of large language models (LLMs), remains challenging for subjects like math. It is crucial to not only identify plausible distractors but also understand the \emph{error} behind them. In this paper, we introduce \mbox{DiVERT} (\textbf{Di}stractor Generation with \textbf{V}ariational \textbf{E}rrors \textbf{R}epresented as \textbf{T}ext), a novel variational approach that learns an interpretable representation of errors behind distractors in math MCQs. Through experiments on a real-world math MCQ dataset with $1,434$ questions used by hundreds of thousands of students, we show that \mbox{DiVERT}, despite using a base open-source LLM with 7B parameters, outperforms state-of-the-art approaches using GPT-4o on downstream distractor generation. We also conduct a human evaluation with math educators and find that \mbox{DiVERT} leads to error labels that are of comparable quality to human-authored ones. 
\end{abstract}

%
%

\section{Introduction}

\begin{figure*}
\centering
\includegraphics[width=\linewidth]{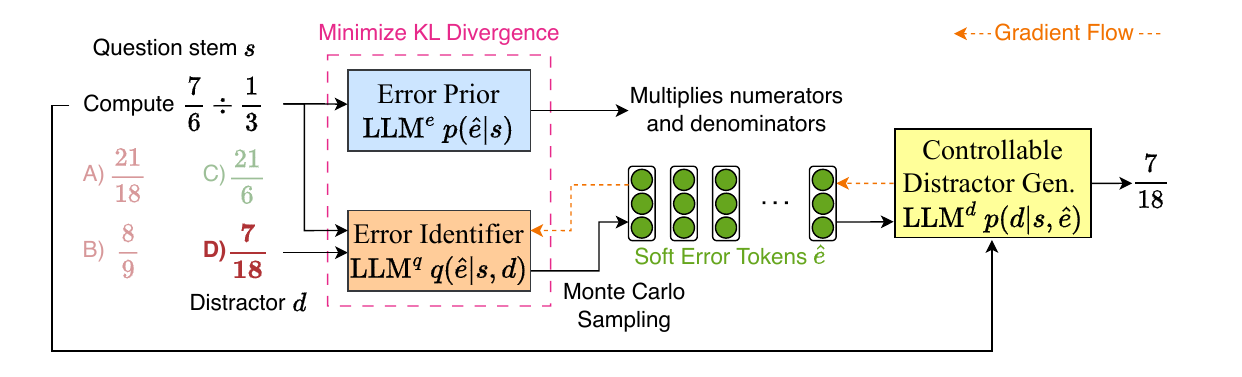}
\caption{Overview of \mbox{DiVERT}'s variational pipeline for error explanation and distractor generation in math MCQs.}
\label{fig:model}
\end{figure*}

Multiple-choice questions (MCQs) are arguably the most common form of questions found in standardized tests. Each MCQ contains a question \emph{stem} that states the context of the questions and the task to be completed, and a series of \emph{options}: a \emph{key}, i.e., the correct answer, embedded among several other \emph{distractors}, i.e., incorrect answers. See Figure~\ref{fig:model} for an example. MCQs are widely used in real-world, large-scale educational/psychological tests and surveys, mainly due to the ease in automated grading \cite{Nitko:96, Airasian:01, Kubiszyn:16}. However, constructing a good set of distractors can be quite challenging: On the one hand, they should correspond to clear errors in the factual recall or reasoning process required by the question's task. On the other hand, these errors should not be too obvious such that no student would select the distractors. Therefore, designing high-quality MCQs that measure specific knowledge components/concepts/skills and more importantly, corresponding distractors good at capturing specific knowledge deficiencies among real students/test takers is very important to the development of high-quality MCQs. 

The typical approach to MCQ distractor development primarily relies on extensive human effort, which can be burdensome for educators, which motivated the development of \emph{automated} approaches. Prior work on automated distractor generation primarily focuses on i) cloze tasks in language learning to assess vocabulary recall or grammatical knowledge and ii) reading comprehension question answering to assess comprehension of a given text or article~\cite{alhazmi2024distractor}. Approaches include using knowledge graphs, encoder-decoder models, and with help from large language models (LLMs). See Section~\ref{sec:rw} for a more detailed review of related work.

For other subjects where common tasks require (possibly complex) reasoning ability, such as math, there exist relatively few approaches to automated MCQ distractor generation. In these domains, generating high-quality distractors is more challenging than in language learning or reading comprehension: distractors need to reflect abstract mathematical misconceptions and/or procedural errors in the mathematical reasoning process. Earlier works either use symbolic, rule-based approaches to generate distractors \cite{tomas2013automatic,prakash2023q}, which have limited generalizability beyond template-based questions, or sample incorrect generations during a math problem solving process as distractors \cite{dave2021math}. More recently, \cite{feng2024exploring,scarlatos2024improving} investigate a wide array of LLM-based approaches for math MCQ generation, particularly focusing on the plausibility of distractors, i.e., how likely is a distractor going to be selected among real students. 

However, to the best of our knowledge, no existing approach has attempted to generate \emph{explanations} of errors underlying MCQ distractors. In assessment scenarios where one's goal is to measure the overall ability of a student/test taker, being able to generate a good set of distractors may be sufficient. However, in real-world educational scenarios where one's goal is to maximize the learning gain of students, being able to \emph{interpret} the cause of error behind a distractor is highly important; this information can be used for student knowledge \emph{diagnosis}, i.e., identifying areas where they lack sufficient knowledge \cite{vanlehn1982bugs} or even pinpoint specific misconceptions they exhibit \cite{wang2021results}. Such diagnosis can be used to provide \emph{feedback} to both teachers, to help them better understand students' learning progress and directly to students in real-time, through intelligent tutoring systems \cite{cogtutor}, online learning platforms \cite{heffernanx2}, or via chatbots powered by LLMs \cite{khanmigo}. Therefore, interpreting the errors behind students selecting specific distractors in math MCQs is important yet challenging, partly due to the mathematical reasoning processes required by these questions.


\subsection{Contributions}

\begin{enumerate}[noitemsep,topsep=0.2pt]
    \item We introduce 
    \mbox{DiVERT}~\footnote{
    Code: \url{https://github.com/umass-ml4ed/divert}
    }(\textbf{Di}stractor Generation with \textbf{V}ariational \textbf{E}rrors \textbf{R}epresented as \textbf{T}ext), a novel variational approach to jointly learn error representations in math MCQs as well as generate distractors corresponding to errors.
    \item We add interpretability to \mbox{DiVERT} by using LLMs to parameterize all distributions in our variational approach. This design enables us to represent and interpret errors as text tokens.
    \item We conduct extensive quantitative and qualitative experiments on a real-world math MCQ dataset. We find that \mbox{DiVERT}, by jointly learning to represent errors and generate distractors, outperforms state-of-the-art approaches on distractor generation.
    \item We conduct a human evaluation with math educators and find that \mbox{DiVERT} leads to error explanations comparable in quality to human-authored ones, and significantly outperforms GPT-4-generated errors despite using a much smaller base LLM with 7B parameters. 
    
\end{enumerate}

%
%

\section{Problem Formulation}

We denote each math MCQ as $Q = \{s, k, f, t, D, E\}$, which contains a set of textual components, including the question stem $s$, the key $k$, (optionally) an explanation of the key $f$, (optionally) question topic/concept tags $t$, and a set of distractors $D$: we denote each distractor itself as $d_i \in D$, with $e_i \in E$ denoting the \emph{error explanation} associated with the distractor, e.g., a misconception.
In this paper, we do not consider MCQs with figures or diagrams since the open-source LLMs we work with cannot process visual input. All of these textual components are sequences of words and math symbols and we denote them as a series of tokens $\{w_1, \ldots, w_L\}$, with $L$ being the length of the sequence. 

We study the task of learning an interpretable space of errors behind distractors in math MCQs. Since it is difficult to quantitatively evaluate error representations, we also study the downstream task of distractor generation \cite{feng2024exploring}. Specifically, our two major goals are: 
\begin{enumerate}[noitemsep,topsep=0.2pt]
    \item \textbf{Error generation:} From a set of real-world math MCQs, learn an error representation space underlying distractors. Possible errors include insufficient knowledge on required skills or exhibiting specific misconceptions.
    \item \textbf{Distractor generation:} For each plausible error, generate corresponding distractor(s), i.e., incorrect answer(s) that an incorrect approach with that error leads to.
\end{enumerate}
We select distractor generation as the downstream task since performance on this task is easily quantifiable; however, there are other meaningful tasks, such as feedback message and tutoring dialogue turn generation, where current methods struggle to implicitly identify student errors \cite{mcnichols2024can, scarlatos2024feedback}, yet knowing what error a student made significantly improves LLMs' performance on these tasks \cite{learnlm}. We leave exploring such tasks for future work. 

We formulate the error generation task as learning a function that outputs a plausible error in an MCQ, given the question stem and attributes including the key and (optionally) explanation and topic tags, i.e.,
$g^\text{err}(s,k,f,t) \rightarrow \hat{e}$. Similarly, we formulate the distractor generation task as learning a function that outputs a distractor that corresponds to a plausible error, i.e.,
$g^\text{dis}(s,k,f,t,\hat{e}) \rightarrow \hat{d}$.

%
%

\section{Methodology}

We now detail our approach to error representation learning and distractor generation in math MCQs. 

\subsection{Variational Approach}

There can be numerous plausible errors among real students for each math MCQ, all of which may lead to different distractors. For example, in the MCQ in Figure~\ref{fig:model} on fraction division, the distractor $\frac{7}{18}$ may result from a student exhibiting the error ``multiplies numerators and denominators'', while the distractor $\frac{8}{9}$ may result from the error ``adds numerators and denominators''. Without explicit error annotations, an error $e$ is a latent variable underlying a distractor $d$. Thus, the observed likelihood of a distractor given a question stem $s$ is
\begin{align}
    p(d|s) = \textstyle\sum_{e \in \mathcal{E}} p(e|s)p(d|s,e),
\end{align}
where $p(e|s)$ is the probability that the error $e$ can be made on question $s$, $p(d|s,e)$ is the probability that $d$ corresponds to the error $e$, and $\mathcal{E}$ is the space of all possible errors, which can be computationally intractable depending on how the errors are represented. While there are many potential representations of the errors, such as discrete categories or continuous latent vectors, we choose to represent each error as a \textit{sequence of textual tokens}, so that the learned underlying representations of distractors are \textit{interpretable} by both humans and language models.

Given this perspective, we naturally use a variational approach to learn an approximation of the large latent error space. Similar to variational auto-encoders \cite{vae}, we jointly learn the latent distribution and maximize the likelihood of the data by maximizing the evidence lower bound (ELBO) on the observed data log-likelihood, given by
\begin{align}
    \label{eq:elbo}
    \log p_{\theta}&(d|s) \geq \text{ELBO}(d|s)\nonumber = \mathcal{L}(\theta,\phi)\\
    &= \mathbb{E}_{q_{\phi}(e|s,d)}[\log p_{\theta_d}(d|s,e)]\nonumber\\
    &- \beta D_\text{KL}(q_{\phi}(e|s,d)\ ||\ p_{\theta_e}(e|s)),
\end{align}
where $\theta_e$, $\theta_d$, and $\phi$ are the parameters of the prior model, distractor likelihood model, and variational model, respectively. The parameter $\beta > 0$ controls the balance between the distractor reconstruction loss and the KL divergence between the approximate posterior and the prior.

To maximize the objective above, we jointly train three models, each parameterized by an LLM:
\begin{enumerate}[noitemsep,topsep=0.2pt]
    \item The \textbf{error prior model} $p_{\theta_e}(e|s)$: This model generates a textual explanation of an error that students can make given the MCQ's stem.
    \item The \textbf{controllable distractor generation model} $p_{\theta_d}(d|s,e)$: This model generates a distractor that is the resulting incorrect answer after making a specific error in this question.
    \item An \textbf{error identifier} model $q_\phi(e|s,d)$: This approximate posterior model generates the error behind a question-distractor pair, and acts as the variational distribution during training.
\end{enumerate}

We note that there can be many choices for the variational distribution $q(\cdot)$; we choose $q(e|s,d)$ since it can be useful in practice by allowing us to recover textual errors from distractors. We leave experimenting with other useful approximate distributions such as $q(e|t)$ or even simply $q(e)$ for future work. Next, we detail how we train all three models in a connected fashion, with the pipeline shown in Figure~\ref{fig:model}.


\subsection{DiVERT}
\label{sec:divert}

As discussed above, to promote interpretability in error representations, we parameterize all distributions in our variational approach using LLMs.  
We use a different LLM for each distribution; each is fine-tuned from the same base LLM using QLoRA~\cite{dettmers2024qlora}. 
We denote them as $\text{LLM}^e$, $\text{LLM}^d$, and $\text{LLM}^q$. Next, we detail a series of novel methods that we introduce to enable \mbox{DiVERT} to effectively perform the tasks of error learning and distractor generation.


\paragraph{Differentiable Learning through Discrete Tokens.}
The ELBO in Equation~\ref{eq:elbo} can easily be approximated by Monte Carlo simulation,  with samples drawn from $q_{\phi}(e|s,d)$ (see Supplementary Material \ref{sec:monte-carlo-approx}). 
However, since we parameterize $q_{\phi}$ with an LLM, these samples are sequences of discrete text tokens. Therefore, we cannot simply use the reparameterization trick to sample from $q_{\phi}$. Using a similar approach to \cite{liu2023boltforsofttokens}, we use soft tokens, i.e., differentiable versions of hard, discrete text tokens during training to enable the flow of gradients. Specifically, for the $k$-th error token $\hat{e}_{k}$ generated by $\text{LLM}^q$, we replace its discrete embedding $\mathbf{v}_{\hat{e}_{k}}$ with $\sum_j p_{k,j} \mathbf{v}_j$, where $p_{k,j}$ is the probability of generating token $j$ at position $k$ from $\text{LLM}^q$ and $\mathbf{v}_j$ is the embedding vector of token $j$ in the vocabulary of $\text{LLM}^q$. This approximation enables us to sample discrete tokens while differentiating through continuous vectors during training. We calculate $p_{k,j} = \operatorname{softmax}(\mathbf{z}_k / \lambda)_j$, where $\mathbf{z}_k$ is the output logit vector from $\text{LLM}^q$ at position $k$ and $\lambda > 0$ is a temperature parameter. 
Following prior work~\cite{jang2017categorical}, we initialize $\lambda$ to $1$ and exponentially anneal it to $0.1$ during training. 
This process means that the soft tokens are smooth when training starts to enable better gradient flow, while later becoming closer to hard tokens.

\paragraph{Initialization with Supervised Fine-tuning.}
Despite LLMs exhibiting better and better mathematical reasoning capabilities, their capacity in understanding errors \cite{sonkar2023deduction} and generating distractors that correspond to errors \cite{feng2024exploring} remains surprisingly poor. Therefore, we solicit a collection of error labels behind question-distractor pairs, $(s,d)$, from math educators. Then, we fine-tune all three LLMs in their respective formats using this annotation data to initialize \mbox{DiVERT}'s three LLM components. Without this step, base LLMs, even ones that perform well on question answering tasks, generate low-quality error explanations that hurt performance, especially during early training stages. In Section~\ref{sec:results}, we demonstrate that using a small portion of error labels for warm-up fine-tuning helps \mbox{DiVERT}'s ability to learn from unlabeled $(s,d)$ pairs.

\paragraph{Q Regularization.}
Since the space of plausible errors can be large, we add a regularization term in our loss function to prevent the approximate posterior distribution $q_{\phi}(e|s,d)$ from deviating too much compared to its initialized version after supervised fine-tuning, $q_{\phi_\text{init}}$. This term, weighted by a balancing parameter $\alpha > 0$, controls the amount of exploration in our error tokens sampled from $q_{\phi}(e|s,d)$ and prevents it from diverging during training. Concretely, the regularization is given by
\begin{align}
    \label{eq:q_reg}
    \mathcal{L}_\text{reg} \!\!=\!\! \textstyle \sum_{k=1}^{|\hat{e}|} \!\! D_\text{KL}(q_{\phi_{\text{init}}}(\hat{e}_k|s,d)||q_{\phi}(\hat{e}_k|s,d)),
\end{align}
with our final training objective to minimize becoming
\begin{align}
    \label{eq:final_objective}
    -\mathcal{L}(\theta,\phi) + \alpha \mathcal{L}_\text{reg}. 
\end{align}
We note that this term is similar to the KL penalty used when training LLMs with reinforcement learning \cite{ouyang2022training}, specifically NLPO \cite{ramamurthy2022reinforcement} that also uses a token-level penalty. However, the key difference is that we directly backpropagate gradients from this loss rather than use it to form a reward.

\paragraph{Overgenerate-and-rank for Errors and Distractors.}
At test time, we first generate a set of $N_e$ errors, denoted as $\hat{E}$, through $\text{LLM}^e$, using diverse beam search \cite{vijayakumar2018diverse} for decoding, to promote diversity among generated errors.
Then, for each generated error $\hat{e} \in \hat{E}$, we generate $N_d$ distractors through $\text{LLM}^d$ using standard beam search since the distractors exhibit much less variation under a specific error. Finally, we rank all $N_e \times N_d$ candidate distractors in $\hat{D}$ by their associated beam scores and select the top-$K$~\cite{ashok-kumar-etal-2023-improving}.

%
%

\section{Experimental Evaluation}
\label{sec:expts}
In this section, we detail our experiments on a real-world math MCQ dataset. For quantitative evaluation, we compare \mbox{DiVERT} against state-of-the-art approaches and strong baselines on the task of distractor generation. For qualitative evaluation, we perform a human evaluation of the generated error explanations with math educators.


\subsection{Dataset Details}
We work with a real-world math MCQ dataset from the Eedi learning platform\footnote{\url{https://eedi.com/us}}, containing $1,434$ MCQs written in English, each with a set of $3$ distractors. We collect error labels from middle school math teachers for each question-distractor pair, explaining why a student may select that distractor. 
The questions are designed primarily to assess students aged between $10$ to $13$, on $41$ unique subtopics, including ``Basic Arithmetic'', ``Fractions'', and ``Solving Equations''. We divide the dataset into train-val-test splits by \emph{questions} to ensure no overlap in the MCQ stem across the splits, resulting in roughly a $72\%$-$16\%$-$12\%$ split over question-distractor pairs. See Supplementary Material~\ref{sec:dataset_details} for statistics, and Supplementary Material~\ref{sec:mcq_example} for MCQ examples.


\subsection{Metrics}

\paragraph{Error Evaluation.}
The open-ended and mathematical nature of errors makes automated text similarity metrics like ROUGE-L F1~\cite{lin-2004-rouge} and BERTScore F1~\cite{bert-score} unsuitable. 
Therefore, we conduct a \textbf{human evaluation} of generated errors, which we detail in Section~\ref{sec:human_eval_errors}. For completeness, we report error evaluation on automated metrics in Supplementary Material~\ref{sec:app_error_eval}.

\paragraph{Distractor Evaluation.}
Following prior work on automated distractor generation \cite{feng2024exploring}, we use alignment-based metrics to measure how well the $K$ generated distractors align with ground-truth human-authored ones.  
The first metric, \textbf{Exact match} ($h_e$), measures whether all three human-authored distractors in $D$ are matched exactly by some subset of the generated distractors $\hat{D}$. Similarly, \textbf{Partial match} ($h_p$) measures whether at least one generated distractor matches human-authored ones. Concretely, these binary metrics are defined as  $h_e(D, \hat{D})=1$ if $D \subseteq \hat{D}$, and $0$ otherwise, while $h_p(D, \hat{D})=1$ if $\hat{D} \cap D \neq \emptyset$, and $0$ otherwise.
The third, continuous metric, \textbf{Proportional match} ($h_n$), measures the portion of human-authored distractors that match generated ones, defined as $h_n(D, \hat{D}) = |\hat{D} \cap D| / 3$.
We compute all metrics averaged across all MCQs in the test set and report percentages. We vary $K \in \{3, 10\}$, similar to the setup for metrics such as MAP@K. The Proportional match metric is most important since it is more robust than the other two.


\subsection{Baselines}

We compare our variational approach, \mbox{DiVERT}, to state-of-the-art distractor generation approaches as well as several strong baselines, outlined below.

\begin{table*}[ht]
\small
\centering
\begin{tabular}{p{0.33\linewidth}p{0.08\linewidth}p{0.08\linewidth}p{0.08\linewidth}|p{0.08\linewidth}p{0.08\linewidth}p{0.08\linewidth}}

\toprule

\multirow{3}{*}{Model} & \multicolumn{3}{c}{K=3} &  \multicolumn{3}{c}{K=10}\\
\cmidrule{2-7}
& Exact@3 & Partial@3 & Prop@3 & Exact@10 & Partial@10 & Prop@10\\

\midrule

\rowcolor{gray!21} \multicolumn{7}{c}{Proprietary Base LLM, GPT-4o}\\
GPT-4o Zero-shot CoT~\cite{feng2024exploring} & $6.22_{\pm2.18}$ & $69.14_{\pm3.97}$ & $35.27_{\pm1.74}$ & $19.47_{\pm3.30}$ & $78.66_{\pm3.71}$ & $50.00_{\pm1.96}$\\ 
GPT-4o kNN~\cite{feng2024exploring} & $\textbf{21.28}_{\pm3.43}$ & $\textbf{78.42}_{\pm5.81}$ & $\textbf{49.63}_{\pm3.88}$ & $33.47_{\pm3.48}$ & $85.14_{\pm4.77}$ & $60.19_{\pm3.87}$\\ 

\rowcolor{gray!21} \multicolumn{7}{c}{Open-source Base LLM, MetaMath-Mistral 7B}\\
DisSearch-D & $13.74_{\pm2.86}$ & $74.13_{\pm4.52}$ & $41.76_{\pm0.50}$ & $34.12_{\pm2.33}$ & $86.17_{\pm4.91}$ & $61.41_{\pm2.68}$\\ 
DisSearch-ED CoT & $\underline{14.11}_{\pm1.22}$ & $73.18_{\pm3.81}$ & $42.14_{\pm1.52}$ & $\underline{36.21}_{\pm1.21}$ & $\underline{86.77}_{\pm3.80}$ & $\underline{62.83}_{\pm2.52}$\\ 
DisSearch-ED CoT Pipeline & $13.53_{\pm1.74}$ & $73.63_{\pm4.05}$ & $41.26_{\pm1.87}$ & $32.97_{\pm3.48}$ & $86.23_{\pm3.88}$ & $60.42_{\pm2.80}$\\ 
DiVERT (ours) & $13.37_{\pm1.70}$ & $\underline{76.33}_{\pm4.33}$ & $\underline{42.87}_{\pm2.45}$ & $\textbf{37.00}_{\pm3.29}$ & $\textbf{87.26}_{\pm4.29}$ & $\textbf{63.24}_{\pm3.37}$\\ 

\bottomrule
\end{tabular}
\caption{Cross-validation performance on distractor generation for all approaches across all metrics. \mbox{DiVERT}, using an open-source base LLM with $7$B parameters, outperforms all baselines and performs on par with or better than the much larger and proprietary GPT-4o. Best performance is in \textbf{bold} and second best is \underline{underlined}.}
\label{tab:results}
\end{table*}








\paragraph{Prompting Baselines.}
We compare \mbox{DiVERT} to two prompting-based approaches proposed in \cite{feng2024exploring} using the state-of-the-art LLM GPT-4o \cite{gpt-4o}. In contrast to our approach, these approaches generate all distractors in one pass, without considering errors behind them. The first is \textbf{kNN}, their best-performing approach, where we use the 3 most similar MCQs in the training set as in-context examples and teacher-written errors as chain-of-thought reasoning to aid distractor generation. We only use questions that have errors for all distractors as examples. The second is \textbf{CoT}, where no in-context examples are given but the model is prompted to generate errors as chain-of-thought~\cite{wei2022chain} reasoning before each distractor. 
We note that~\citet{feng2024exploring} use teacher-written feedback instead of errors as CoT, but we use errors for a fairer comparison to our method.
We make other small updates to the prompts to include question tags and to generate either 3 or 10 distractors per question. All our prompts are shown in Supplementary Material~\ref{sec:prompts}. In Supplementary Material~\ref{sec:more_baselines}, we show results from other baselines in~\citet{feng2024exploring} on our data. We also perform a preliminary, small-scale comparison with the recent OpenAI o1 model~\cite{o1}, which reportedly performs better than GPT-4o on mathematical reasoning tasks.

\paragraph{Fine-Tuning Baselines.}
We introduce three strong fine-tuning baselines for distractor generation. The first is \textbf{DisSearch-D}, where we fine-tune the base LLM to directly generate a distractor from the question stem, i.e., training $p(d|s)$ on all question-distractor pairs.  
The second is \textbf{DisSearch-ED CoT}, where we fine-tune the base LLM to generate the error first, as chain-of-thought~\cite{wei2022chain} reasoning, followed by the distractor, i.e., training $p(e,d|s)$ on all question-distractor pairs.  
At inference time, we also use beam search to match the overgenerate-and-rank setup used in \mbox{DiVERT}. 
The third is \textbf{DisSearch-ED CoT Pipeline}, where we use the fine-tuned $p(d|s,e)$ and $p(e|s)$ models without variational training and test with the same two-step pipeline as \mbox{DiVERT}.


\subsection{Experimental Setup}
\label{sec:exp_setup}

For the prompting baselines, we use the same implementation and hyperparameters as the public code repository in~\citet{feng2024exploring} for a fair comparison. We use the latest base LLM from the GPT-4 family, GPT-4o (as of June 13, 2024), instead of GPT-3.5, to further strengthen their performance. 
For \mbox{DiVERT}, we use MetaMath-Mistral $7$B~\cite{yu2023metamath} as our base LLM since it is one of the best-performing LLMs in the $7$B family on mathematical reasoning; we found that it outperforms other open-source LLMs with similar size on our tasks.
See Supplementary Material~\ref{sec:params} for detailed parameter settings. 
At test time, we overgenerate $N_e=10$ errors via $\text{LLM}^e$ using diverse beam search~\cite{vijayakumar2018diverse} and $N_d=10$ distractors through $\text{LLM}^d$ using standard beam search. Among the set of $N_e \times N_d=100$ error-distractor pairs we select the top-$K \in \{3,10\}$. 
For DisSearch-ED CoT, we use standard beam search with $100$ beams for a fair comparison with \mbox{DiVERT}. DisSearch-D performs better with $10$ beams so we report its performance with $10$ beams. For the primary results in Table~\ref{tab:results}, we perform a 5-fold cross-validation, rotating the test set over the dataset, and report the average and standard deviation of metrics across methods. For all other experiments, we evaluate on a single fold to reduce training time.

%
%

\section{Results, Analysis, and Discussion}
\label{sec:results}

In this section, we quantitatively evaluate the quality of generated distractors, qualitatively evaluate both errors and distractors, perform error analyses on failed cases, and conduct an ablation study.

\subsection{Quantitative Evaluation}

\paragraph{\mbox{DiVERT} performs comparably or better than GPT-4o and outperforms baselines.} 
Table~\ref{tab:results} shows downstream distractor generation performance for all approaches on all evaluation metrics. Our variational approach, DiVERT, using an open-source base LLM with 7B parameters, performs on par with GPT-4o, especially on the robust Proportional@$10$ metric, where it outperforms GPT-4o by a wide margin. GPT-4o kNN performs best on $K=3$ evaluation metrics, which is not surprising since prior work~\cite{feng2024exploring} found that the kNN approach exploits in-context examples in the training set that have the same underlying structure as the target MCQ, different only in named entities and numerical values. Therefore, all it needs to do is to follow patterns and generate three distractors correspondingly without understanding errors. However, it does not perform as well on $K=10$ metrics since it often fails to go beyond the top three errors in MCQs that have numerous plausible errors. In contrast, \mbox{DiVERT} performs well by training on error labels to acquire an understanding of plausible mathematical errors. 
Under the same base LLM, \mbox{DiVERT} performs better than baselines on almost all metrics, which highlights the importance of its variational approach, since sampling from the $\text{LLM}^q$ model during training encourages exploration of the error space and improves model robustness.

\paragraph{Errors as ``chain-of-thought'' improve performance.}
Comparing among baselines, we see that training on human-authored error explanations improves distractor generation performance. This result can easily be explained by error labels serving as valuable chain-of-thought~\cite{wei2022chain} supervision during training.

\begin{figure}
\centering
\includegraphics[width=1\linewidth]{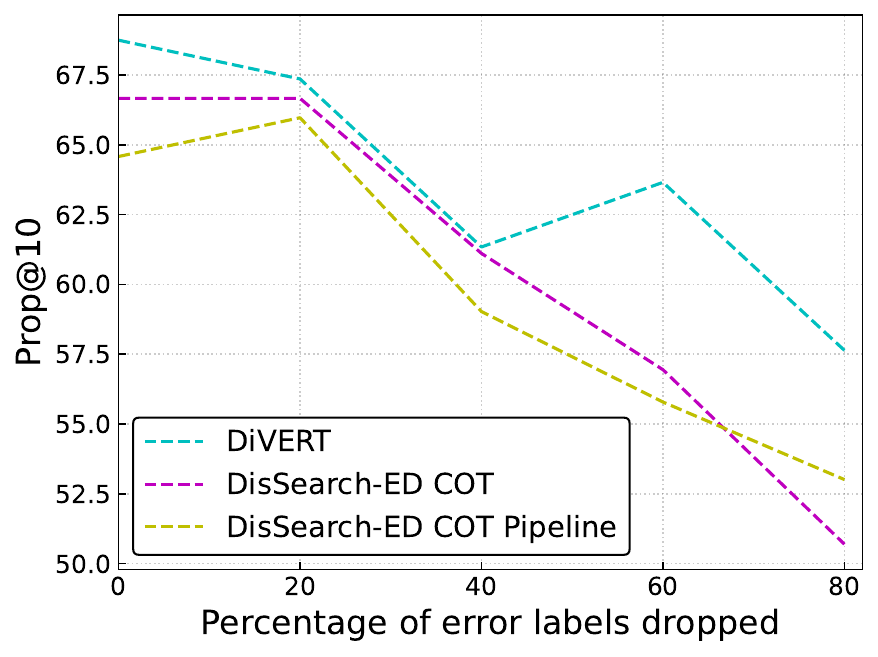}
\caption{Distractor generation performance with increasing percentages of error labels dropped (unused in training). \mbox{DiVERT} outperforms baselines, especially when only a small number of error labels are used.} 
\label{fig:vae_graph}
\end{figure}

\paragraph{DiVERT works even with a small number of error labels.} 
Since soliciting error labels behind question-distractor pairs from math educators is time-consuming, we investigate \mbox{DiVERT}'s reliance on the amount of error label data. Figure~\ref{fig:vae_graph} shows \mbox{DiVERT}'s distractor generation performance using different portions of the available error labels. We see that on the most stable metric, Prop@$10$, \mbox{DiVERT} performs better than baselines, especially when the majority of error labels are dropped: the performance gap is widest when less than half of the training error labels are used for training. This result further highlights the importance of the self-exploration training setup in \mbox{DiVERT}'s variational approach at going beyond human-authored labels and linking different error types across questions.

\paragraph{Ablation study.}

\begin{table}
\small
\centering
\begin{tabular}{p{0.45\linewidth}p{0.1\linewidth}p{0.1\linewidth}p{0.1\linewidth}}

\toprule
Model & Exact & Partial & Prop\\

\midrule
DiVERT & $\textbf{42.36}$ & $\textbf{91.67}$ & $\textbf{68.75}$ \\ 
$-$ variational training & $35.41$ & $90.27$ & $64.81$ \\
$-$ train $q_\phi$ (freeze $q_\phi$) & $37.50$ & $91.66$ & $66.20$ \\ 
$-$ $q_\phi$ regularization & $40.97$ & $89.58$ & $66.20$ \\ 
$-$ temperature annealing & $41.66$ & $89.58$ & $67.36$ \\ 

\bottomrule
\end{tabular}
\caption{Ablation study of \mbox{DiVERT} on distractor generation performance on $K=10$ metrics on a single fold.}
\label{tab:results_ablation}
\end{table}

Table~\ref{tab:results_ablation} shows the results of the ablation study on the more robust @$10$ metrics on a single fold. We see that the variational training process is crucial and removing it results in a large drop in performance (see Supplementary Material~\ref{sec:q_increase_samples_graph}). We also see that not training $q_{\phi}$ and using its fine-tuned initialized version $q_{\phi_{init}}$ also drops performance, which highlights the importance of sampling from $q_{\phi}$ and exploring the error space of math MCQs. Regularizing $q_{\phi}$ and using temperature annealing in the soft token approximation are also effective at making training more robust.


\subsection{Qualitative Evaluation}

Table~\ref{tab:qual_ex_divert} shows generated errors and corresponding distractors for an MCQ stem in the test set. 
\mbox{DiVERT} and GPT-4o kNN each generate a diverse set of plausible errors and corresponding distractors.
However, GPT-4o Zero-shot CoT prioritizes error diversity and generates errors unlikely to be made by real students like ``Confusing the concept of multiples with that of prime factors''.
The best fine-tuning baseline, DisSearch-ED CoT, generates generic error descriptions like ``Does not understand the term multiple'', which is not specific to the question's context. 

\begin{table}
\small
\centering
\begin{tabular}{p{0.76\linewidth}p{0.14\linewidth}}

\toprule

\multicolumn{2}{c}{Q: What is the lowest common multiple of $12$ and $15$?}\\

\midrule
Error & Distractor\\

\midrule
\rowcolor{gray!21} \multicolumn{2}{c}{Ground-Truth}\\

\midrule
Believes finding the product of two numbers gives their lowest common multiple. & $180$\\
Believes the largest number in a set of numbers is always their lowest common multiple. & $15$\\
Confuses factors and multiples. & $3$\\

\midrule
\rowcolor{gray!21} \multicolumn{2}{c}{DiVERT}\\

\midrule
Believes they can find the lowest common multiple by adding the numbers together. & $27$\\
Confuses factors and multiples. & $3$\\
When asked for the lowest common multiple, thinks they can just give any multiple of one of the numbers. & $15$\\

\midrule
\rowcolor{gray!21} \multicolumn{2}{c}{GPT-4o kNN}\\
\midrule
Confuses the product of the two numbers with the LCM. & 180\\
Incorrectly identifies a common multiple that is not the lowest. & 120\\
Incorrectly adds the two numbers. & 27\\

\midrule
\rowcolor{gray!21} \multicolumn{2}{c}{GPT-4o Zero-shot CoT}\\
\midrule
Confusing the concept of the lowest common multiple with the greatest common divisor. & 3\\
Incorrectly adding the two numbers instead of finding the lowest common multiple. & 27\\
Confusing the concept of multiples with that of prime factors. & 5\\


\midrule
\rowcolor{gray!21} \multicolumn{2}{c}{DisSearch-ED CoT}\\
\midrule
Confuses factors and multiples.  & $3$\\
Does not understand the term multiple. & $15$\\
Identifies a common multiple but not the lowest common multiple. & $75$\\

\bottomrule
\end{tabular}
\caption{Examples of errors and corresponding distractors generated by different approaches for a test MCQ.}
\label{tab:qual_ex_divert}
\end{table}

\paragraph{Failure Pattern Analysis.}
We now investigate failure patterns in the generated errors and corresponding distractors from \mbox{DiVERT}. Table~\ref{tab:qual_error_analyses} shows a representative example MCQ in the test set. We observe that a majority of generated errors from $p(e|s)$ are mathematically valid, with high diversity, but some are less likely among real students such as ``Subtracts instead of divides'', matching the observation made in prior work~\cite{feng2024exploring}. 
By far, the most frequent failure pattern we observe is on the controllable distractor generation model $p(d|s,e)$, where the generated distractor is not faithful/consistent to the error.  
In this example, for the error ``When dividing a fraction by an integer, divides the denominator by the integer'', the generated distractor is $\frac{6}{1}$ rather than the consistent $\frac{6}{3}$. The reverse can also occasionally happen when a distractor is plausible but the error is not. 
This observation highlights that although effectively learning good error representations and capable of identifying what errors can be made in a math MCQ, the biggest limitation of \mbox{DiVERT} is in its inability to enforce consistency in the downstream distractor generation model $p(d|s,e)$. This limitation suggests a major direction for future work, possibly by exploring the use of an error-distractor consistency penalty in the training objective.

%
%

\section{Human Evaluation of Errors}
\label{sec:human_eval_errors}


\subsection{Evaluation Setup}

While automated distractor evaluation is well-defined since there is ground-truth, it is more challenging to automatically evaluate errors. Reference-based evaluation may penalize errors that faithfully reflect the mathematical error behind a distractor but are semantically different than the ground truth, or conversely, reward errors that are invalid but semantically similar to the ground truth. To address this challenge, we conduct a human evaluation to measure the quality of generated and human-authored errors. We recruit two experienced math teachers who have extensive expertise in designing math MCQs as annotators. We randomly select $20$ questions from the test set on a diverse set of topics, and for each question, show annotators the ground-truth, human-authored errors, and generated errors from both \mbox{DiVERT} and GPT-4o, for a total of $180$ errors. For \mbox{DiVERT}, we select the top 3 distractors from the $p(e|s)$ model using diverse beam search to promote error diversity. For GPT-4o, we select the errors generated during CoT distractor generation. We randomize the order of errors shown to annotators and do not tell them how each error is generated. We instruct annotators to rate each error on a 5-point Likert scale, with 5 being the best, depending on whether an error is relevant to the question, mathematically sound, specific, conceptual, and likely to be made by a real student. See Supplementary Material \ref{sec:human-eval-instructions} for the exact instructions given to annotators.

\begin{table}
\small
\centering
\begin{tabular}{p{0.75\linewidth}p{0.1\linewidth}}

\toprule

\multicolumn{2}{c}{Question stem: Calculate: $\frac{6}{9} \div 3$}\\ 
\midrule
Error & Distractor\\
\midrule

\multicolumn{2}{c}{\textcolor{OliveGreen}{Plausible} error, \textcolor{OliveGreen}{plausible} and 
\textcolor{OliveGreen}{consistent} distractor.}\\
\midrule
When dividing a fraction by an integer, divides both the numerator and denominator by the integer. & $\frac{2}{3}$\\

\midrule

\multicolumn{2}{c}{\textcolor{OliveGreen}{Plausible} error, \textcolor{OliveGreen}{plausible} but 
\textcolor{CornellRed}{inconsistent} distractor.}\\
\midrule
When dividing a fraction by an integer, divides the denominator by the integer. & $\frac{6}{1}$\\

\midrule
\multicolumn{2}{c}{\textcolor{CornellRed}{Implausible} error, \textcolor{OliveGreen}{plausible} but 
\textcolor{CornellRed}{inconsistent} distractor.}\\
\midrule
Divided by the denominator instead of the numerator. & $\frac{1}{3}$\\

\midrule
\multicolumn{2}{c}{\textcolor{CornellRed}{Implausible} error, \textcolor{CornellRed}{implausible} and 
\textcolor{CornellRed}{inconsistent} distractor.}\\
\midrule
Subtracts instead of divides. & $\frac{3}{6}$\\

\bottomrule
\end{tabular}
\caption{Qualitative error analyses of generated errors and corresponding distractors from \mbox{DiVERT} on an example MCQ stem from the test set.}
\label{tab:qual_error_analyses}
\end{table}


\subsection{Results}

\begin{table}[t]
    \small
    \centering
    \begin{tabular}{cccc}
        \toprule
         & Human & DiVERT & GPT-4o \\
        \midrule
        Rating & $3.23 \pm 1.28$ & $3.07 \pm 1.39$ & $2.56 \pm 1.25$ \\
        \bottomrule
    \end{tabular}
    \caption{Average error quality rated by math teachers. Human and \mbox{DiVERT} errors are similar in quality, and both are better than GPT-4o with statistical significance.}
    \label{tab:human-eval-results}
\end{table}

Table~\ref{tab:human-eval-results} shows the average and standard deviation of annotators' ratings on errors that were human-authored and LLM-generated, by both \mbox{DiVERT} and GPT-4o. We find that the quality of \mbox{DiVERT}'s errors are close to human ones, with no statistically significant difference between them using a Two-Sample $t$-Test ($p=0.36$). This result suggests that \mbox{DiVERT} retains the human-level quality of errors it is trained on during fine-tuning. 
Moreover, we find that both human and \mbox{DiVERT}-generated errors are better than GPT-4o ones, with statistical significance ($p<0.01$). This result is promising since our base LLM, MetaMath-Mistral 7B, is orders of magnitude smaller than GPT-4o. 
Qualitatively, we find that GPT-4o's errors are often not what real students are likely to make, and even occasionally confuse the correct solution approach with an error. This result shows that even state-of-the-art LLMs are not able to anticipate student errors, and that training on human-authored labels is likely necessary for LLMs to accurately diagnose student errors. 
Finally, we note that overall, the error labels vary a lot in terms of quality and score lower than expected. The Pearson correlation coefficient between our annotators' ratings is also only $0.33$, indicating low-to-moderate agreement. This result is due to many errors deemed to be unlikely to be made by students by annotators, rather than being mathematically incorrect. It is likely that even for humans, anticipating errors made by students is a challenging task, which suggests that future work should focus on understanding the nature of errors made by real students and their causes.

%
%

\section{Related Work}
\label{sec:rw}

For automated distractor generation in language learning and reading comprehension, prior works have explored ranking candidate distractors based on semantic similarity to the key and using knowledge graphs \cite{susanti2018automatic, stasaski2017multiple,alsubait2014generating}. More recent works use an end-to-end pipeline for distractor generation, which lead to longer and higher-quality distractors \cite{distractor-gen-ui, qiu2020automatic, shuai2023qdg, xie2021diverse, gao2019generating}, also leveraging open-source LLMs such as BERT and T5 \cite{kalpakchi2021bert, chiang2022cdgp, rodriguez2022end, qu2024unsupervised, wang-etal-2023-distractor}; these approaches are similar to the baselines we use in this work. Other works prompt state-of-the-art proprietary LLMs such as ChatGPT and GPT-4 to generate distractors, with carefully crafted prompts, for a wider range of subjects including math and computer science \cite{tran2023generating,bitew2023distractor,feng2024exploring}. Evaluating the quality of distractors, however, remains challenging: \cite{moore2024automatic} proposes a series of features that can be used to evaluate the quality of distractors, although those features are mostly about surface semantics and do not apply to subjects like math that require deeper reasoning. Another line of recent work in \cite{feng2024exploring,scarlatos2024improving} proposes to measure the quality of distractors through how likely they are going to be selected by real students, which requires training another model to predict student option selection behavior. However, most existing MCQ datasets do not come with such student response information. 

At a high level, an alternative to our approach of representing errors as text is to learn \emph{latent} error representations \cite{li2020optimus,tu2022adavae}, i.e., characterizing errors as a latent stochastic error vector, possibly in the text embedding space. However, in our experiments, we found this approach to be uninterpretable and ineffective, significantly reducing distractor generation performance, possibly because errors behind distractors in math MCQs correspond to complex reasoning processes and are harder to capture than different user writing styles. Our approach in improving the diversity of generated errors also bears some resemblance to the works in \cite{wen2022equal,wang2023elaboration}, although the nature of our task is more complex than dialogue generation and question answering.

%
%

\section{Conclusions and Future Work}

In this paper, we proposed \mbox{DiVERT}, a novel variational approach to jointly learn an interpretable, textual representation of errors in math MCQs and how to generate distractors that correspond to these errors. On a real-world math MCQ dataset, we showed that \mbox{DiVERT} results in better downstream distractor generation performance over state-of-the-art approaches. We also conducted a human evaluation with math educators and found that \mbox{DiVERT} leads to errors as interpretable as those provided by humans, and outperforms GPT-4o, despite using a much smaller base LLM. 

There are many avenues for future work. First, we plan to apply the learned error representations to another downstream task, feedback generation, and explore whether our approach leads to a more accurate student profile/model, which will in turn result in higher quality feedback in math tutoring chatbots. Second, we plan to test the across-topic generalizability of our approach and investigate whether the error representations can generalize to previously unseen math topics. Third, we hope to develop new, LLM-based metrics to evaluate the mathematical validity of an error and the equivalence between two error explanations, to alleviate the need for human evaluation of error quality.


\section*{Acknowledgements}
The authors would like to thank the annotators and Jaewook Lee, Hunter McNichols, Hasnain Heickal, and Nancy Otero for helpful discussions. We also thank the anonymous reviewers for their helpful comments. This work is partially supported by Schmidt Futures under the learning engineering virtual institute (LEVI) initiative and the NSF under grants 2118706, 2237676, and 2341948.

%
%

\section*{Limitations}

We identify several technical and practical limitations of our work. First, the main limitation of \mbox{DiVERT} is its tendency to generate distractors that do not always correspond with the preceding generated error. However, we note that all baselines (including GPT-4) also exhibit this behavior, and we plan on addressing this limitation as a line of future work. Second, we note that \mbox{DiVERT} requires a set of human-authored errors for initialization through fine-tuning. This process of labeling distractors with textual errors can be time-consuming and difficult to scale, although we observe that \mbox{DiVERT} retains most of its ability when only a small subset of the data is labeled. Third, we only experiment with one dataset, since to the best of our knowledge there are no similar distractor datasets with labeled textual errors. Finally, we do not perform a human evaluation on the quality of the distractors themselves due to limited resources.

%
%

\section*{Ethical Considerations}

Our goal in this work is to develop a system that can automatically create distractors for assessments or practice problems, and do so in an explainable way so that the distractors can be easily verified by human educators. We hope that such systems will save educators time on content creation, allowing them to spend more resources on personal student interactions. However, there is a concern that such systems could replace human educator jobs, which is a shared concern across most domains with AI applications. We note that while there is little risk for bias in the creation of mathematical distractors, the use of textual errors introduces the possibility of generating biased text by inheriting tendencies from the base LLM. Another risk of automatically generating distractors is that lower-quality learning content compared to human-authored content could lead to negative student learning outcomes. Because of these reasons, we recommend that generated errors and distractors be reviewed by experts before being deployed to real students.

\bibliography{custom}

\clearpage
\appendix

%
%

\section{Monte Carlo Approximation of ELBO Training Objective}
\label{sec:monte-carlo-approx}
Since exact inference of the posterior distribution is computationally intractable, $q_{\phi}(e|s,d)$ is an approximation of the true posterior. We perform approximate inference by maximizing the evidence lower bound (ELBO) on the observed data log-likelihood, given in Equation~\ref{eq:elbo}. The ELBO in Equation~\ref{eq:elbo} can easily be approximated by Monte Carlo simulation, with samples drawn from $q_{\phi}(e|s,d)$, as shown below:
\begin{align*}
    \mathbb{E}&_{q_{\phi}(e|s,d)}[\log p_{\theta_d}(d|s,e)]\nonumber\\
    & - \beta D_\text{KL}(q_{\phi}(e|s,d)\ ||\ p_{\theta_e}(e|s))\nonumber\\
    = & \mathbb{E}_{q_{\phi}(e|s,d)}[\log p_{\theta_d}(d|s,e)]\nonumber\nonumber\\
    & - \beta \mathbb{E}_{q_{\phi}(e|s,d)}[\log q_{\phi}(e|s,d)\ -\ \log p_{\theta_e}(e|s)]\nonumber\\
    \approx & \textstyle \sum_{\hat{e} \sim q_{\phi}(e|s,d)} \log p_{\theta_d}(d|s,\hat{e}) - \beta \log q_{\phi}(\hat{e}|s,d)\nonumber\\
    & + \beta \log p_{\theta_e}(\hat{e}|s)
\end{align*}

\section{Additional Baselines}
\label{sec:more_baselines}

In Table \ref{tab:results_extra}, we show results on distractor generation for additional baselines from \cite{feng2024exploring}, including kNN using feedback instead of errors and the rule-based method, where GPT-4o first selects errors from a pool before generating distractors. We also run the CoT baseline using the OpenAI o1 model \cite{o1}. To reduce costs, we collect these additional results on a single fold of the data, and show the other GPT-4o baselines and \mbox{DiVERT} performance on this fold for reference.

We first observe that CoT with o1 only slightly outperforms CoT with GPT-4o, improving Prop@10 by 1.62. This indicates that the long internal reasoning generated by o1 does not increase alignment with teacher-written distractors, though it may reduce arithmetic or logical errors resulting in the slight increase. We next observe that the Rule-Based method is slightly better than CoT, though still significantly worse than kNN, consistent with findings in \cite{feng2024exploring}. Finally, we observe there is little difference when using feedback instead of errors in kNN. This result validates our choice of using errors as textual reasoning before generating distractors, and also indicates that the exact form of the reasoning may not be significant in the context of generating aligned distractors.

\begin{table*}[ht]
\small
\centering
\begin{tabular}{p{0.33\linewidth}p{0.08\linewidth}p{0.08\linewidth}p{0.08\linewidth}|p{0.08\linewidth}p{0.08\linewidth}p{0.08\linewidth}}

\toprule

\multirow{3}{*}{Model} & \multicolumn{3}{c}{K=3} &  \multicolumn{3}{c}{K=10}\\
\cmidrule{2-7}
& Exact@3 & Partial@3 & Prop@3 & Exact@10 & Partial@10 & Prop@10\\

\midrule

\rowcolor{gray!21} \multicolumn{7}{c}{Proprietary Base LLM, OpenAI o1}\\
OpenAI o1 Zero-shot CoT & $4.17$ & $72.22$ & $36.11$ & $25.69$ & $77.78$ & $52.08$\\

\rowcolor{gray!21} \multicolumn{7}{c}{Proprietary Base LLM, GPT-4o}\\
GPT-4o Zero-shot CoT~\cite{feng2024exploring} & $2.78$ & $68.75$ & $33.33$ & $20.83$ & $81.25$ & $50.46$\\ 
GPT-4o Rule-Based~\cite{feng2024exploring} & $4.17$ & $67.36$ & $35.42$ & $26.39$ & $77.78$ & $53.70$\\ 
GPT-4o kNN Feedback~\cite{feng2024exploring} & $\textbf{25.69}$ & $\underline{83.33}$ & $\textbf{55.09}$ & $\underline{36.81}$ & $\underline{88.89}$ & $\underline{63.89}$\\ 
GPT-4o kNN Errors & $\underline{23.61}$ & $\textbf{85.42}$ & $\underline{52.78}$ & $\underline{36.81}$ & $\underline{88.89}$ & $63.66$\\ 

\rowcolor{gray!21} \multicolumn{7}{c}{Open-source Base LLM, MetaMath-Mistral 7B}\\
DiVERT (ours) & $13.19$ & $81.25$ & $46.06$ & $\textbf{42.36}$ & $\textbf{91.67}$ & $\textbf{68.75}$\\ 

\bottomrule
\end{tabular}
\caption{Single fold performance on distractor generation for additional baselines and reference methods. Best performance is in \textbf{bold} and second best is \underline{underlined}.}
\label{tab:results_extra}
\end{table*}

%
%

\section{Variational Training Ablation: Q Model Helps \mbox{DiVERT} Learn a Robust Error Space}
\label{sec:q_increase_samples_graph}

\begin{figure}
\centering
\includegraphics[width=1\linewidth]{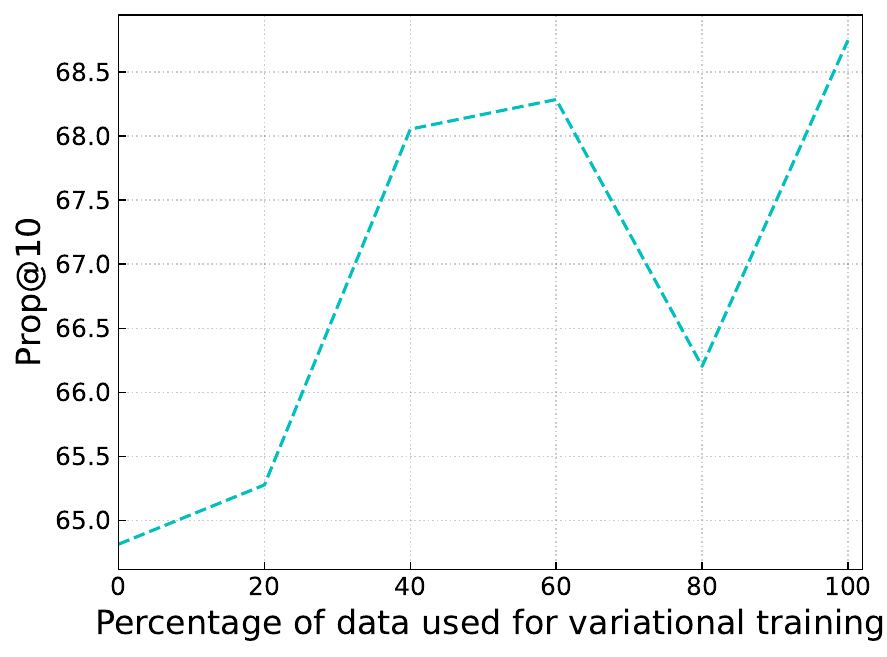}
\caption{Distractor generation Prop@10 performance with an increasing percentage of data used for variational training. Sampling errors from $q_{\phi}$ on all train question-distractor pairs performs best.} 
\label{fig:q_increase_samples_graph}
\end{figure}

\begin{table*}
\small
\centering
\begin{tabular}{p{2.35cm}||c|c|c|c||c|c|c|c||c|c|c|c}

\toprule

\multirow{4}{*}{Math MCQ Dataset}
& \multicolumn{4}{c||}{Train} & \multicolumn{4}{c||}{Validation} & \multicolumn{4}{c}{Test}\\
\cmidrule{2-13}
& \multicolumn{4}{c||}{$2570$ question-distractor pairs} & \multicolumn{4}{c||}{$599$ question-distractor pairs} & \multicolumn{4}{c}{$432$ question-distractor pairs}\\
\cmidrule{2-13}
& $\mu$ & $\sigma$ & Min & Max & $\mu$ & $\sigma$ & Min & Max & $\mu$ & $\sigma$ & Min & Max\\

\midrule

\# tokens/Q stem 
& $43.2$ & $30.9$ & $5$ & $224$
& $42.3$ & $31.5$ & $5$ & $174$
& $35.1$ & $27.9$ & $6$ & $164$\\
\# tokens/solution
& $66.0$ & $35.3$ & $12$ & $234$
& $67.0$ & $39.0$ & $15$ & $266$
& $59.3$ & $29.7$ & $11$ & $164$\\
\# tokens/key
& $9.3$ & $11.5$ & $1$ & $189$
& $9.5$ & $14.6$ & $1$ & $175$
& $7.5$ & $4.7$ & $1$ & $44$\\
\# tokens/distractor
& $9.4$ & $11.6$ & $1$ & $184$
& $9.7$ & $15.3$ & $1$ & $176$
& $5.1$ & $3.7$ & $1$ & $31$\\
\# tokens/error
& $14.3$ & $6.0$ & $4$ & $42$
& $14.0$ & $6.1$ & $5$ & $36$
& $13.8$ & $5.5$ & $5$ & $39$\\

\bottomrule
\end{tabular}
\caption{Statistics of the real-world math MCQ dataset which contains $1,434$ MCQs across $41$ unique subtopics.}
\label{tab:math_mcq_dataset_stats}
\end{table*}

In real-world educational scenarios, there are multiple errors plausible among real students for the same MCQ stem. \mbox{DiVERT}, can learn a robust error space, by leveraging sampled errors from $q_{\phi}(e|s,d)$ around the original human-authored ones during variational training. As shown in Figure~\ref{fig:q_increase_samples_graph}, sampling errors from $q_{\phi}$ on a higher number of question-distractor pairs in the train set, leads to a more robust learned error space, with better downstream distractor generation performance.

%
%

\section{Real-world Math MCQ Dataset Details}
\label{sec:dataset_details}

We divide the real-world math MCQ dataset into train-val-test splits by \emph{questions} to ensure no overlap in the MCQ stem across the splits, resulting in roughly a $72\%$-$16\%$-$12\%$ split over question-distractor pairs. Table~\ref{tab:math_mcq_dataset_stats} shows detailed statistics of the dataset. 
Each of the $1434$ MCQs has $3$ distractors leading to $1434 \cdot 3 = 4302$ question-distractor pairs. Among these pairs, $3601$ pairs have human annotated error descriptions; other pairs are not labeled due to the non-mathematical nature of errors, e.g., careless slipping, reading the question incorrectly, etc.
There are $1231$ unique error descriptions. For the test set, we only keep questions containing human annotated error descriptions for all three associated distractors, leading to $144$ questions with $144 \cdot 3 = 432$ question-distractor pairs.
We manually checked a random subset of the data and found no personally identifying information or offensive content.

%
%
\section{Automated Error Evaluation}
\label{sec:app_error_eval}

\subsection{Metrics}

The open-ended and mathematical nature of errors makes automated text similarity metrics like ROUGE-L F1~\cite{lin-2004-rouge} and BERTScore F1~\cite{bert-score} unsuitable. 
Therefore, we conduct a \textbf{human evaluation} of generated errors, which we detail in Section~\ref{sec:human_eval_errors}. For completeness, we report error evaluation on automated metrics below.

We evaluate generated errors on two key aspects: 1) similarity with ground-truth, human-authored errors $E$ with $|E|=3$, and 2) diversity. 
We select the best $3$ errors generated for each MCQ, i.e., $|\hat{E}|=3$. We compute both recall, which evaluates how well the generated errors recover actual human-authored errors, and precision, which evaluates how accurate the generated errors are with respect to the human-authored errors.
Concretely, we measure recall by
\begin{align*}
    \text{sim}_r^{h}(E,\hat{E}) = \textstyle\sum_{e \in E} \text{max}_{\hat{e} \in \hat{E}} (h(e,\hat{e})) / |E|
\end{align*}
and precision by
\begin{align*}
    \text{sim}_p^{h}(E,\hat{E}) = \textstyle\sum_{\hat{e} \in \hat{E}} \text{max}_{e \in E} (h(\hat{e},e)) / |\hat{E}|,
\end{align*}
where $h$ denotes the choice of the textual similarity metric. We use traditional textual similarity metrics including \textbf{ROUGE-L F1}~\cite{lin-2004-rouge} and \textbf{cosine similarity} using the pre-trained SBERT encoder MPNet~\cite{song2020mpnet}, as well as recent metrics like \textbf{BERTScore F1}~\cite{bert-score}.

For \textbf{diversity}, following~\cite{padmakumar2024does,shaib2024standardizing}, we report the complement of the homogenization score of a set of errors $E$ as
\begin{align*}
    \text{div}^{h}(E) = 1 - \textstyle \sum_{e_1 \neq e_2 \in E} h(e_1,e_2) / |E\times E|,
\end{align*}
where $h$ denotes the choice of the textual similarity metric. We report diversity for both human-authored errors $E$ and predicted errors $\hat{E}$, averaged across all test MCQs.

\subsection{Baselines and Results}

\begin{table*}
\small
\centering
\begin{tabular}{p{0.18\linewidth}p{0.04\linewidth}p{0.04\linewidth}p{0.04\linewidth}p{0.04\linewidth}|p{0.04\linewidth}p{0.04\linewidth}p{0.04\linewidth}p{0.04\linewidth}|p{0.04\linewidth}p{0.04\linewidth}p{0.04\linewidth}p{0.04\linewidth}}

\toprule

\multirow{3}{*}{Model} & \multicolumn{4}{c}{ROUGE-L F1} &  \multicolumn{4}{c}{BERTScore F1} & \multicolumn{4}{c}{Cosine Similarity}\\
\cmidrule{2-13}
& Precis. & Recall & F1 & Div. & Precis. & Recall & F1 & Div. & Precis. & Recall & F1 & Div.\\

\midrule

\rowcolor{gray!21} \multicolumn{13}{c}{Proprietary Base LLM, GPT-4o}\\

GPT-4o Zero-shot CoT & $0.251$ & $0.272$ & $0.261$ & $0.718$ & $0.632$ & $0.635$ & $0.633$ & $0.293$ & $0.602$ & $0.607$ & $0.605$ & $0.376$\\ 

\rowcolor{gray!21} \multicolumn{13}{c}{Open-source Base LLM, MetaMath-Mistral 7B}\\

ErrorSearch-E & $\underline{0.498}$ & $\textbf{0.597}$ & $\underline{0.543}$ & $\underline{0.781}$ & $\underline{0.741}$ & $\textbf{0.786}$ & $\underline{0.763}$ & $\underline{0.368}$ & $\underline{0.698}$ & $\textbf{0.759}$ & $\textbf{0.727}$ & $\underline{0.475}$\\ 

DisSearch-ED CoT & $\textbf{0.595}$ & $0.526$ & $\textbf{0.558}$ & $0.448$ & $\textbf{0.786}$ & $0.751$ & $\textbf{0.768}$ & $0.207$ & $\textbf{0.735}$ & $0.702$ & $\underline{0.718}$ & $0.278$\\

DiVERT $p(e|s)$ (ours) & $0.479$ & $\underline{0.576}$ & $0.523$ & $\textbf{0.786}$ & $0.732$ & $\underline{0.775}$ & $0.753$ & $\textbf{0.372}$ & $0.680$ & $\underline{0.746}$ & $0.711$ & $\textbf{0.487}$\\  

\bottomrule
\end{tabular}
\caption{Performance on automated error evaluation for all error-based approaches across all metrics. Best performance is in \textbf{bold} and second best is \underline{underlined}.}
\label{tab:results_error_eval}
\end{table*}

We introduce a new fine-tuning baseline for error generation, \textbf{ErrorSearch-E}, where we fine-tune the base LLM to directly generate an error from the question stem, i.e., training $p(e|s)$ on all error-question pairs. For a fair comparison, we generate errors from the $p(e|s)$ model of \mbox{DiVERT} in a standalone fashion. We use diverse beam search decoding~\cite{vijayakumar2018diverse} to generate errors from both models.
We also compare with errors generated from GPT-4o Zero-shot CoT, as well as the finetuning baseline DisSearch-ED CoT, both of which generate errors followed by distractors.

Table~\ref{tab:results_error_eval} shows error generation performance for all error-based approaches on all evaluation metrics. 
As a reference for the diversity of predicted errors shown, the diversity of ground-truth, human-written errors is $0.574$, $0.300$, and $0.349$, for the choice of the similarity metric as ROUGE-L F1, cosine similarity, and BERTScore F1, respectively.
The finetuning baseline ErrorSearch-E imitates the ground-truth human-written error distribution and performs best on recall performance across all textual similarity metrics. DisSearch-ED CoT performs best on precision performance across all textual similarity metrics. However, the same beam search decoding helping precision, leads to a drop in the performance of DisSearch-ED CoT on recall and diversity. GPT-4o Zero-shot CoT exhibits good error diversity, but as expected performs poorly on textual similarity metrics, with the zero-shot errors generated not matching the distribution of human-written errors. The $p(e|s)$ model from our variational method, \mbox{DiVERT}, generates errors with the highest diversity. This diversity also leads to a slight drop in overall F1 performance across textual similarity metrics. This result is not surprising since by design, during the variational training of \mbox{DiVERT}, the $p(e|s)$ model aligns with the entropy model $q_{\phi}(e|s,d)$, which is encouraged to generate error samples around human-written errors to learn a robust error space representation, leading to better downstream distractor generation performance.

We note that reference-based evaluation may penalize errors that faithfully reflect the mathematical error behind a distractor but are semantically different than the ground truth, or conversely, reward errors that are invalid but semantically similar to the ground truth. Therefore, we conduct a \textbf{human evaluation} of generated errors, which we detail in Section~\ref{sec:human_eval_errors}.

%
%

\section{Experimental Setup}
\label{sec:params}

As detailed in Section~\ref{sec:divert}, we finetune all three LLMs, $\text{LLM}^e$, $\text{LLM}^d$, and $\text{LLM}^q$, using the collection of error label annotations behind question-distractor $(s,d)$ pairs obtained from middle school math teachers, to initialize $p_{\theta_e}(e|s)$, $p_{\theta_d}(d|s,e)$, and $q_{\phi}(e|s,d)$, respectively. We use the AdamW~\cite{loshchilov2018decoupled} optimizer with a batch size of $32$, a learning rate of $2e$-$5$, and perform gradient clipping for training stability. We use the Parameter Efficient Fine-Tuning (PEFT) library from HuggingFace~\cite{wolf-etal-2020-transformers} to load the base LLM, MetaMath-Mistral $7$B, and train via low-rank adaptation (LoRA)~\cite{hu2022lora} ($\text{LoRA }\alpha=256, \text{LoRA }r=128, \text{LoRA dropout}=0.05$) using $8$-bit quantization~\cite{dettmers2024qlora}. We fine-tune for $5$ epochs with early stopping on the validation set on a single NVIDIA A$100$ $80$GB GPU, with each epoch taking up to $35$ minutes. We follow the same training setup for our fine-tuning baselines, DisSearch-D, DisSearch-ED, and DisSearch-ED CoT Pipeline.

After initialization, we perform \mbox{DiVERT} training for $\text{LLM}^e$, $\text{LLM}^d$, and $\text{LLM}^q$ using the same QLoRA setup as above. We use Monte Carlo simulation to approximate the ELBO in Equation~\ref{eq:elbo} with $4$ error samples drawn from $q_{\phi}(e|s,d)$. We use AdamW with a learning rate of $5e$-$6$, matching the learning rate in MetaMath finetuning~\cite{yu2023metamath}, and perform gradient clipping for training stability. A single batch contains $16$ question-distractor pairs, each having $4$ Monte Carlo samples, for an effective batch size of $64$. We set $\beta$ in Equation~\ref{eq:elbo} to $0.1$, following prior work~\cite{aritra_mnss} to upweight the reconstruction loss. We set $\alpha$ in Equation~\ref{eq:final_objective} to $0.95$ to upweight the ELBO compared to the Q regularization loss. We train for $1$ epoch on a single NVIDIA A$100$ $80$GB GPU, which takes up to $9$ hours.
Wherever possible, we use standard hyperparameters and do not do extensive parameter tuning like a grid search. Due to high computational and Open AI API cost, we report performance on one run of our \mbox{DiVERT} model and baselines.
For metrics, for ROUGE we use the rouge-score library with Porter stemmer enabled, and for BERTScore we use the bert-score library with microsoft/deberta-xlarge-mnli as the underlying model.
We additionally note that we used GitHub Copilot minimally in the writing of our code. All software we use in the development of this work is open source. We are consistent with the terms and intended use of all software and with the OpenAI API.

%
%

\section{Example MCQs from Real-world Math MCQ Dataset}
\label{sec:mcq_example}
We show example MCQs from the dataset in Table~\ref{tab:mcq_example}.

\begin{table*}
\small
\centering
\begin{tabular}{p{0.15\linewidth} p{0.73\linewidth}}

\toprule
\rowcolor{gray!21} Question stem & James starts counting from $-2$, adding one each time. What is the $5$th number he says?\\
\midrule
Topic & Adding and Subtracting Negative Numbers\\
Concept & Count forwards starting from a negative integer including through zero\\
Solution & $2$\\
Correct answer & Starting on $-2$, we add one each time, moving up towards and then beyond $0$ until we reach the $5$th number, which is $2$.\\
\midrule
Distractor 1 & $6$\\
Error 1 & Counts on by $2$, when asked to count forward in steps of $1$\\
\midrule
Distractor 2 & $3$\\
Error 2 & Counts on from the wrong number\\
\midrule
Distractor 3 & $-6$\\
Error 3 & Counts on from the wrong number'\\

\midrule
\rowcolor{gray!21} Question stem & $7^2 = ?$\\
\midrule
Topic & Squares, Cubes, etc\\
Concept & Calculate the square of a number\\
Solution & $7^2 = 7\times 7 = 49$\\
Correct answer & $49$\\
\midrule
Distractor 1 & $14$\\
Error 1 & Mixes up squaring and multiplying by $2$ or doubling\\
\midrule
Distractor 2 & $72$\\
Error 2 & Reads a power as a normal digit\\
\midrule
Distractor 3 & $77$\\
Error 3 & Mixes up squaring with repeating a digit\\

\midrule
\rowcolor{gray!21} Question stem & What is the highest common factor of $8$ and $28$?\\
\midrule
Topic & Factors and Highest Common Factor\\
Concept & Identify the Highest Common Factor of two numbers\\
Solution & $8$ has factors $1$, $2$, $4$ and $8$ and $28$ has factors $1$, $2$, $4$, $7$, $14$ and $28$. The highest factor common to both is $4$.\\
Correct answer & $4$\\
\midrule
Distractor 1 & $28$\\
Error 1 & Believes the largest number in a set of numbers is always their highest common factor\\
\midrule
Distractor 2 & $8$\\
Error 2 & Believes the smallest number in a set of numbers is always their highest common factor\\
\midrule
Distractor 3 & $2$\\
Error 3 & Identifies a common factor but not the highest common factor\\
\bottomrule

\end{tabular}
\caption{Example MCQs from the real-world math MCQ dataset.}
\label{tab:mcq_example}
\end{table*}

%
%
\section{Prompts}
\label{sec:prompts}

\subsection{Prompts for Base LLMs in \mbox{DiVERT}}

We show all prompts used for the base LLMs in \mbox{DiVERT}, the error prior model $p(e|s)$ parameterized by $\text{LLM}^e$ in Table~\ref{tab:prompt-divert-llm-e}, the controllable distractor generation model $p(d|s,e)$ parameterized by $\text{LLM}^d$ in Table~\ref{tab:prompt-divert-llm-d}, and the error identifier model $p(e|s,d)$ parameterized by $\text{LLM}^q$ in Table~\ref{tab:prompt-divert-llm-q}.

\begin{table*}
    \small
    \centering
    \begin{tabular}{p{0.95\linewidth}}
    \toprule
    A teacher assigns the following math question to a class of middle school students.\\
    The question is: <question stem>\\
    The question topic is: <topic>\\
    The question concept is: <concept>\\
    The solution is: <worked out solution>\\
    The correct answer is: <answer>\\
    A possible error made by a student is:\\
    \bottomrule
    \end{tabular}
    \caption{Prompt for error prior model $p(e|s)$ parameterized by $\text{LLM}^e$ in \mbox{DiVERT}.}
    \label{tab:prompt-divert-llm-e}
\end{table*}

\begin{table*}
    \small
    \centering
    \begin{tabular}{p{0.95\linewidth}}
    \toprule
    A teacher assigns the following math question to a class of middle school students.\\
    The question is: <question stem>\\
    The question topic is: <topic>\\
    The question concept is: <concept>\\
    The solution is: <worked out solution>\\
    The correct answer is: <answer>\\
    The error made by the student is: <error>\\
    The incorrect answer given by the student is:\\
    \bottomrule
    \end{tabular}
    \caption{Prompt for controllable distractor generation model $p(d|s,e)$ parameterized by $\text{LLM}^d$ in \mbox{DiVERT}.}
    \label{tab:prompt-divert-llm-d}
\end{table*}

\begin{table*}
    \small
    \centering
    \begin{tabular}{p{0.95\linewidth}}
    \toprule
    A teacher assigns the following math question to a class of middle school students.\\
    The question is: <question stem>\\
    The question topic is: <topic>\\
    The question concept is: <concept>\\
    The solution is: <worked out solution>\\
    The correct answer is: <answer>\\
    The incorrect answer given by the student is: <distractor>\\
    The error made by the student is:\\
    \bottomrule
    \end{tabular}
    \caption{Prompt for error identifier model $q(e|s,d)$ parameterized by $\text{LLM}^q$ in \mbox{DiVERT}.}
    \label{tab:prompt-divert-llm-q}
\end{table*}

\subsection{Prompts for Prompting-based Baselines}
\label{sec:prompts-prompting-baselines}
We show all prompts used for prompting-based baselines, GPT-4o Zero-shot CoT in Table~\ref{tab:prompt-gpt-4o-cot}, and GPT-4o kNN in Table~\ref{tab:prompt-gpt-4o-knn}.

\begin{table*}
    \small
    \centering
    \begin{tabular}{p{0.95\linewidth}}
    \toprule
    You are given the following math question along with the correct answer and explanation. Please use the following template to give <n> alternative incorrect answers to be used as multiple-choice options in a multiple-choice exam. Prior to the incorrect answer, provide the underlying error corresponding to that incorrect answer. These errors should be conceptual in nature and should not refer to numbers, variables, or names in the question.\newline
[Template]\newline
Distractor1 Error:\newline
Distractor1:\newline
...\newline
Distractor<n> Error:\newline
Distractor<n>:\newline

Question: <question>\newline
Topic: <topic>\newline
Concept: <concept>\newline
Explanation: <worked out solution>\newline
Answer: <answer>\\
    \bottomrule
    \end{tabular}
    \caption{Prompt for GPT-4o Zero-Shot CoT.}
    \label{tab:prompt-gpt-4o-cot}
\end{table*}

\begin{table*}
    \small
    \centering
    \begin{tabular}{p{0.95\linewidth}}
    \toprule
    You will be given a math question along with the correct answer and explanation. You will be also provided with several example questions that include incorrect distractor answers. Please generate <n> incorrect distractor answers for the current question to be used as multiple-choice options in a multiple-choice exam.\newline
[Template]\newline
Distractor1 Error:\newline
Distractor1:\newline
...\newline
Distractor<n> Error:\newline
Distractor<n>:\newline

<selected examples>\newline

Question: <question>\newline
Topic: <topic>\newline
Concept: <concept>\newline
Explanation: <worked out solution>\newline
Answer: <answer>\\
    \bottomrule
    \end{tabular}
    \caption{Prompt for GPT-4o kNN.}
    \label{tab:prompt-gpt-4o-knn}
\end{table*}

%
%

\section{Human Evaluation Details}

We received IRB approval for our human evaluation of error quality. Our evaluators were volunteers contacted through a research partner and were not compensated monetarily. They were made aware that their annotations would be used in scientific research in AI. We provide the instructions given to them for evaluating errors in Supplementary Material~\ref{sec:human-eval-instructions}.

\subsection{Human Evaluation Instructions}
\label{sec:human-eval-instructions}

For each error, provide a rating between 1 and 5 (inclusive) in the “rating” column, where 1 is the worst and 5 is the best.

Please use the following criteria for evaluating errors:
\begin{itemize}
    \item Relevant: The error should be applicable to the current question and the way it is solved.
    \item Correct: The error should be mathematically sound and concrete.
    \item Specific: The error should be specific to the question’s topic not be too generic.
    \item Conceptual: The error should be conceptual in nature, such that it could be applied to other similar questions.
    \item Plausible: The error should be likely to be made by some (or many) real students.
    \item Overall Rating: The rating you give should reflect the overall quality of the error across all the above criteria; you may deem that some criteria are more important than others depending on the context, so use your best judgment.
\end{itemize}

\end{document}